\documentclass{article}

\PassOptionsToPackage{numbers, compress}{natbib}
\usepackage{pifont}

\usepackage{makecell}

\usepackage{algorithmic}
\usepackage{microtype}
\usepackage{multirow} 
\usepackage{graphicx}
\usepackage{subcaption}
\usepackage{booktabs} % for professional tables
\usepackage{bm}
\usepackage{hyperref}

\usepackage[preprint]{neurips_2026}
\usepackage{bbm}

\usepackage[utf8]{inputenc} % allow utf-8 input
\usepackage[T1]{fontenc}    % use 8-bit T1 fonts
\usepackage{hyperref}       % hyperlinks
\usepackage{url}            % simple URL typesetting
\usepackage{booktabs}       % professional-quality tables
\usepackage{amsfonts}       % blackboard math symbols
\usepackage{nicefrac}       % compact symbols for 1/2, etc.
\usepackage{microtype}      % microtypography
\usepackage{xcolor}         % colors
\usepackage{algorithm}
\usepackage{algorithmic}
\usepackage{multirow}   
\usepackage{amsthm}
\usepackage{amssymb}
\usepackage{graphicx}
\usepackage{amsmath}
\usepackage{wrapfig}
\usepackage{array} 
\usepackage[table]{xcolor}
\usepackage{subcaption}  
\usepackage{enumitem}
\hypersetup{hidelinks}
\definecolor{lightblue}{RGB}{220,230,241}
\usepackage[table]{xcolor}

\definecolor{lightblue}{RGB}{225,240,255}
\usepackage{amsmath}
\usepackage{amssymb}
\usepackage{mathtools}
\usepackage{amsthm}
\usepackage{enumitem}

\usepackage[capitalize,noabbrev]{cleveref}

\theoremstyle{plain}

\theoremstyle{definition}

\theoremstyle{remark}

\newcommand{\methodname}{PrismFlow}
\usepackage[textsize=tiny]{todonotes}
\title{
PrismFlow: Residual Dynamics for Flow Matching in Time-Series Generation}

\author{
\textbf{Junru Zhang}$^{1}$\quad
\textbf{Lang Feng}$^{2}$\quad
\textbf{Jinbo Wang}$^{1}$\quad
\textbf{Xu Guo}$^{2}$\quad
\textbf{Yucheng Wang}$^{3}$ \\
\textbf{Han Yu}$^{2}$\quad
\textbf{Min Wu}$^{3}$\thanks{Corresponding authors.}\quad
\textbf{Yabo Dong}$^{1}$\footnotemark[1]\quad
\textbf{Duanqing Xu}$^{1}$ \\
$^{1}$Zhejiang University, China \\
$^{2}$Nanyang Technological University, Singapore \\
$^{3}$I2R, Agency for Science, Technology and Research (A*STAR), Singapore \\
}

\begin{document}

\maketitle
\begin{abstract}
Generating high-quality time-series data is challenging because real-world signals often exhibit multimodal patterns and multiscale dynamics, including oscillations and high-frequency variations. Flow Matching (FM) offers an efficient alternative to diffusion models, but practical implementations typically rely on a single finite-capacity global vector-field estimator. In such heterogeneous temporal distributions, distinct regimes may pass through nearby flow states while requiring incompatible conditional velocities. A monolithic estimator trained with the standard $\ell_2$ velocity-matching objective may therefore learn an overly smoothed approximation of the local transport field. This estimator-level smoothing can attenuate branch-specific dynamics, leading to spectral distortion and poor mode coverage.
To address this, we propose \methodname{}, a new FM method with Koopman-inspired dynamical experts. Each expert learns residual corrections in a latent space where local nonlinear temporal evolution can be approximated by linear transitions. We further propose a confidence-aware Winner-Take-All (WTA) objective that updates only the expert best aligned with each sample while masking gradients to the others, encouraging mode-specific specialization. During sampling, the selected expert adds a residual dynamical correction to the global transport field, preserving FM stability while recovering fine-grained and high-frequency temporal structures. Across various benchmarks, \methodname{} effectively mitigates the spectral contraction in standard FM and achieves state-of-the-art performance, with a 15.6\% gain in Context-FID and a 38.6\% improvement in Discriminative Score, while remaining robust in low-data settings and effective for forecasting and imputation.
\end{abstract}
\section{Introduction}
Time-series data underpin decision-making in domains such as healthcare \cite{kaushik2020ai,zhang2025diffusion}, finance \cite{huang2024generative,chitsazdual,liu2024diffusion}, and environmental monitoring \cite{he2024ltcr, coletta2023constrained,wu2024effective}. However, the acquisition of high-fidelity signals is frequently restricted by stringent privacy regulations and prohibitive costs \cite{farayola2024data,gonen2025time}. This scarcity necessitates generative models capable of synthesizing sequences that are not only statistically consistent but also preserve the underlying temporal evolution of real-world phenomena.

Recent advances in time-series generation have shifted from adversarial frameworks \cite{goodfellow2014generative,yoon2019time} toward simulation-free continuous transport methods, most notably Flow Matching (FM) \cite{lipman2022flow,albergo2023stochastic}. While FM offers superior stability and sampling efficiency compared to diffusion models \cite{ho2020denoising,ho2022classifier}, its practical implementation typically relies on a \textit{single global vector-field estimator} to approximate the transport dynamics \cite{schusterbauer2025diff2flow}. Although the exact FM transport field is well-defined in principle, learning it from finite data with a monolithic estimator introduces statistical and representational challenges. In multi-modal temporal scenarios, such as those with varying frequencies or transient responses, heterogeneous temporal patterns may occupy nearby regions of the flow state space while requiring distinct sequence evolutions. Under the standard $\ell_2$ velocity-matching objective, the estimator is therefore encouraged to predict the conditional mean of incompatible target velocities, which tends to smooth out mode-specific time-series patterns. As a result, rich temporal structure can collapse into an overly smooth learned flow with constrained spectral diversity, leading to \textit{mode collapse} \cite{arora2017generalization,bang2018improved,abbahaddou2025a,eide2020sample, pan2022unigan}. This degrades local trajectory fidelity and restricts the expressive capacity of flow-based generators by limiting their ability to reproduce diverse dynamical patterns.

To mitigate this  mode collapse, we propose \methodname{},  a new FM method that corrects the global transport field using a bank of Koopman-inspired dynamical experts. Drawing on Koopman operator theory~\citep{koopman1931hamiltonian, lan2013linearization}, our method maps nonlinear temporal evolutions into a latent space where they can be modeled as local linear transitions. Rather than forcing a single learned estimator to capture all dynamics, we treat generation as a \emph{routing} problem: at each intermediate state, the model dynamically identifies the local temporal pattern and assigns responsibility to the expert best suited to model that specific evolution. To ensure these experts learn distinct modes, we introduce a Winner-Take-All (WTA) training objective with competitive selection. Unlike standard mixtures with soft gating~\cite{shazeer2017outrageously,jacobs1991adaptive}, which can reintroduce averaging effects, our approach enforces clearer expert specialization. For a given state, all experts predict candidate residual velocities, but only the expert with the best confidence-aware WTA score receives the specialization update, while gradients to the others are masked. This mechanism reduces the averaging of incompatible modes, which is the primary cause of regression-to-the-mean behavior in the learned monolithic estimator. During sampling, \methodname{} integrates the selected expert as a \textit{residual dynamical correction} term added to the global transport field, thus preserving the stability of the global flow while allowing the model to recover diverse spectral components across multiple scales.

Our main contributions are summarized as follows:

\begin{itemize}[leftmargin=0.4cm,topsep=-1pt,itemsep=-0.25pt]
    \item We characterize velocity averaging in learned FM models as a key bottleneck for time-series generation. Our analysis shows that single-field estimators can induce conditional-mean behavior, leading to spectral distortion and mode collapse in generated time series.
    \item We propose PrismFlow, a novel method that mitigates {mode collapse} in time-series generation. By integrating Koopman-inspired experts as {residual dynamical correction} terms to the global flow, we capture multi-scale dynamics that standard monolithic estimators often fail to represent.
    \item We propose training these experts with a WTA competitive selection rule. By assigning each flow state to a single expert and masking gradients to the others, this objective reduces regression-to-the-mean behavior under the standard $\ell_2$ loss and promotes clear mode-specific specialization.
    \item Empirical evaluations of PrismFlow demonstrate strong performance across diverse time-series generation tasks. \methodname{} effectively recovers diverse modes, improving Context-FID by 15.6\% and Discriminative Score by 38.6\%, while remaining robust in low-data settings and high-fidelity for forecasting and imputation.
\end{itemize}

\section{Related Work}

\textbf{Time-Series Generation.}
Time-series generation has progressed from early adversarial \cite{mogren2016c,esteban2017real,yoon2019time,xu2020cot,jeha2022psa} and latent-variable \cite{desai2021timevae,qian2021latent} models toward iterative refinement methods that deliver higher fidelity.
Representative baselines such as TimeGAN \cite{yoon2019time} and TimeVAE \cite{desai2021timevae} capture global temporal dependencies, but they often suffer from training instability and tend to blur local, phase-sensitive dynamics.
More recently, diffusion-based models \cite{zhang2024diverse,yang2024survey,kong2020diffwave,yuan2024diffusion}, including Diffwave \cite{kong2020diffwave} and Diffusion-TS \cite{yuan2024diffusion}, have achieved strong synthesis quality by casting generation as progressive denoising.
SDformer \cite{chen2024sdformer} further explores discrete sequence modeling with a large-parameter diffusion model.
While diffusion models provide impressive fidelity and distributional alignment, their many-step inference remains computationally expensive, limiting real-time and large-scale deployment.

\textbf{Flow Matching.}
Flow Matching (FM) is a simulation-free framework for training continuous normalizing flows, combining stable objectives with efficient ordinary differential equation (ODE)-based sampling \cite{lipman2022flow,albergo2023stochastic}.
By regressing time-dependent velocity fields along predefined probability paths, FM avoids expensive trajectory simulations during training and enables deterministic generation via standard solvers.
This efficiency has yielded strong results in image synthesis \cite{esser2024scaling}, video generation \cite{chen2025goku}, stable neural ODE dynamics \cite{tamir2024conditional}, time-series foundation modeling \cite{ge2025t2s, liu2025sundial}, and probabilistic forecasting \cite{kollovieh2024flow,kim2025sequence}.
Recently, TimeMCL \cite{cortes2025winner} introduced a multiple-choice learning approach for forecasting diverse futures.
Complementary to such output-level diversity methods, our work focuses on estimator-level averaging in monolithic FM implementations.
When heterogeneous temporal modes pass through nearby flow states, the standard $\ell_2$ objective may drive a single estimator toward the conditional mean of incompatible velocities, yielding over-smoothed trajectories and reduced spectral diversity. Rather than treating this as a flaw of the exact FM transport field, we address the practical limitation of its finite-sample approximation by augmenting the global estimator with dynamically routed Koopman-inspired residual experts.

\section{Preliminaries}

\textbf{Problem Setting.}
Let $x \in \mathcal{X} = \mathbb{R}^{S \times D}$ denote a multivariate time series with $S$ temporal steps and $D$ channels.
Each sequence $x$ is drawn from an unknown data distribution $q(x)$ over $\mathcal{X}$.
The goal of generative modeling is to learn a parameterized distribution $p_\theta$ that approximates $q$, enabling the synthesis of sequences $\hat{x}$ that preserve both the statistical properties and temporal dynamics of real-world data.

\textbf{Flow Matching.} \label{sec:prelim}
Flow Matching (FM)~\cite{lipman2022flow} is a simulation-free framework for training Continuous Normalizing Flows.
It generates samples by transporting a simple source distribution $p_0$ to the data distribution through a time-dependent vector field
$v_t^\theta : [0,1]\times\mathcal{X}\to\mathcal{X}$.
The transformation is governed by the ODE:
\begin{equation}
\label{eq:fm}
\frac{\mathrm{d}}{\mathrm{d}t}x_t = v_t^\theta(x_t, t),
\qquad x_{t=0} = x_0,
\end{equation}
which induces a probability path $\{p_t\}_{t\in[0,1]}$ with $x_t\sim p_t$ for $x_0\sim p_0$.
We parameterize $v_t^\theta(x_t,t)$ with an encoder-decoder network whose parameters are denoted by $\theta=(\phi_\eta,\phi_\zeta)$.
Following standard Conditional Flow Matching (CFM), we adopt the linear interpolation between $x_0\sim p_0=\mathcal{N}(\mathbf{0},\mathbf{I})$ and $x_1\sim q$:
$
x_t = (1-t)x_0 + t x_1,
$
whose target velocity is constant:
\begin{equation}
\frac{\mathrm{d}}{\mathrm{d}t}x_t = x_1 - x_0.
\end{equation}
The model is trained by minimizing:
\begin{equation} 
\label{equ:cfm}
\mathcal{L}_{\mathrm{CFM}}(\theta)
=
\mathbb{E}_{t,x_0,x_1}
\left[
\left\| v_t^\theta(x_t, t) - (x_1 - x_0) \right\|_2^2
\right],
\end{equation}
where $t \sim \mathcal{U}[0,1]$.
To generate a sample, we draw $x_0 \sim \mathcal{N}(\mathbf{0},\mathbf{I})$ and integrate the learned ODE from $t=0$ to $t=1$:
\begin{equation} 
\label{equ:pf}
x_{t+\Delta t} = x_t + v_t^\theta(x_t, t)\Delta t,
\end{equation}
using a numerical solver to obtain $\hat{x}=x_{t=1}$.

\textbf{Mode Collapse.}~
While the $\ell_2$ objective in Eq.~(\ref{equ:cfm}) is effective, practical single-field estimators can become mean-seeking in multi-modal temporal settings.
In standard FM, when trajectories from different temporal regimes pass through nearby flow states $x_t$, the local target velocity distribution $q(u_t\mid x_t)$ can become highly heterogeneous, where $u_t=x_1-x_0$ denotes the CFM target velocity.
Under the $\ell_2$ loss, a finite-capacity estimator may learn a smoothed approximation around the conditional-average trend,
$
v_t^*(x_t,t)=\mathbb{E}[u_t\mid x_t],
$
rather than effectively preserving branch-specific velocity directions.
When nearby samples correspond to incompatible temporal regimes, this estimator-level averaging can reduce the effective velocity energy,
$
\|\mathbb{E}[u_t \mid x_t]\|_2^2
\le
\mathbb{E}[\|u_t\|_2^2 \mid x_t],
$
and may attenuate transient branches and high-frequency components after ODE integration.
From a Dynamic Mode Decomposition (DMD) perspective~\cite{schmid2010dynamic}, this practical smoothing manifests as spectral contraction, where energy concentrates into a few slowly varying modes while faster or weaker modes vanish.
We study this specific form of mode collapse~\cite{bang2018improved,abbahaddou2025a} in time-series generation, which is further validated by a Gaussian-mixture diagnostic in the Appendix.

\begin{figure*}[t]
\centering
\includegraphics[width=\textwidth]{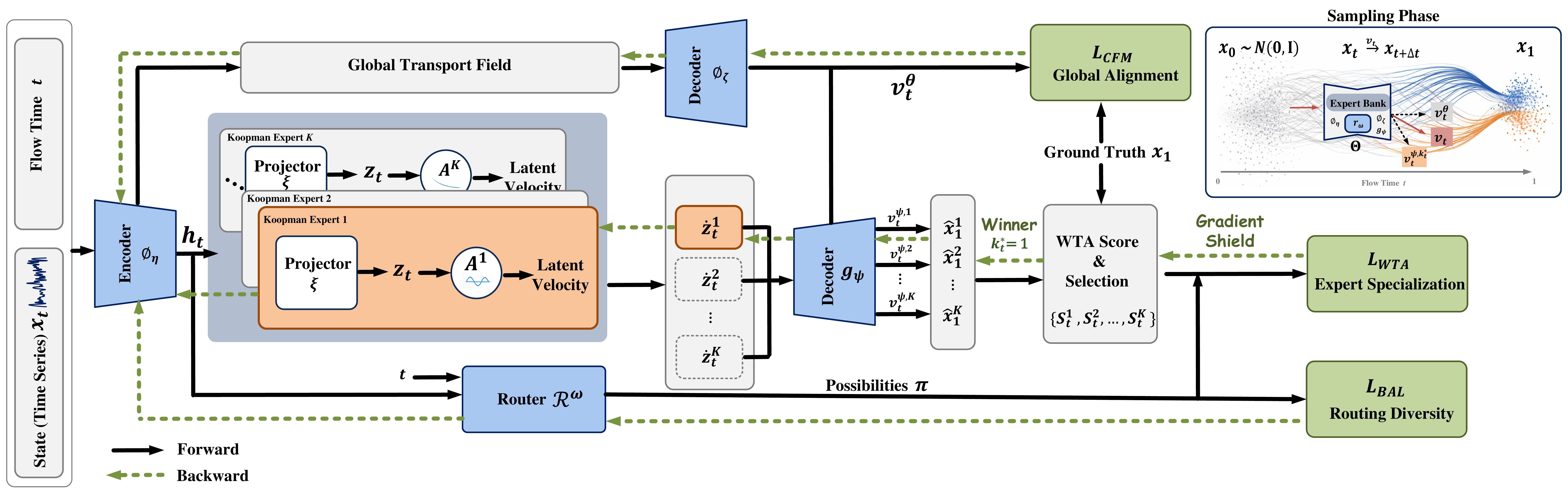}
\caption{Overall architecture of \methodname{}.
Given a flow state \(x_t\), a flow matching backbone predicts a global transport velocity \(v_t^\theta\).
In parallel, a shared encoder and projector map \(x_t\) into a latent Koopman space, where a bank of linear experts models different temporal modes.
Each expert evolves its latent state through a linear velocity, which is decoded by \(g_\psi\) into a candidate residual velocity \(\{v_t^{\psi,k}\}_{k=1}^K\).
A router produces expert probabilities \(\pi_t\), and a WTA score selects the dominant expert \(k_t^*\).
During training, \(\mathcal{L}_{\mathrm{WTA}}\) assigns the specialization signal to the selected expert, \(\mathcal{L}_{\mathrm{CFM}}\) aligns the global transport with the target distribution, and \(\mathcal{L}_{\mathrm{BAL}}\) encourages balanced expert usage.
During sampling, the selected expert provides a residual correction to the global transport field.}
\label{fig:framework}
\vspace{-0.3cm}
\end{figure*}

\section{Methodology}
\label{sec:methodology}

We introduce \methodname{}, a new flow matching method designed to mitigate the estimator-level mean seeking.
As shown in Fig.~\ref{fig:framework}, \methodname{} preserves the global transport process of FM while augmenting it with residual dynamical corrections.
The method contains two core components:
(i) \textit{Mode-Specific Experts}, a bank of Koopman-inspired experts that captures distinct temporal regimes and allows local dynamics to deviate from the global average; and
(ii) \textit{Residual Routing}, a Winner-Take-All (WTA) mechanism that assigns each flow state to a dominant expert.
By learning residual velocities that complement the global estimator, \methodname{} recovers multi-scale temporal structures while retaining the simulation-free efficiency of standard FM.

In the following, we detail the components of \methodname{} by
(\textbf{1}) defining Koopman-based linear experts that provide mode-specific residual dynamics (Sec.~\ref{sec:method_experts});
(\textbf{2}) introducing the WTA routing policy for hard expert selection  (Sec.~\ref{sec:method_wta});
(\textbf{3}) describing the multi-objective training that enforces gradient-masked specialization while balancing expert utilization (Sec.~\ref{sec:method_optimization}); and
(\textbf{4}) presenting the sampling procedure where expert-informed residuals complement the global transport estimator to preserve spectral diversity (Sec.~\ref{sec:method_sampling}).

\subsection{Koopman Experts for Structured Mode-Seeking}
\label{sec:method_experts}

To mitigate estimator-level averaging across incompatible temporal regimes, such as oscillations and sharp transients, we introduce a bank of $K$ experts that capture distinct evolution patterns.
When multiple temporal modes can explain nearby flow states, different experts should take responsibility for different regimes rather than forcing a single estimator to absorb them through a mean-seeking prediction.
To model such structured dynamics efficiently, we adopt Koopman theory~\cite{koopman1931hamiltonian}, which suggests that nonlinear temporal evolution in the original state space can admit an approximately linear description in a latent representation.
Formally, for ${x}^{(s)} \rightarrow {x}^{(s+1)}$, there exist an observable map $f$ and a linear operator $A$ such that
$f({x}^{(s+1)}) = A \circ f({x}^{(s)})$.
We therefore learn a latent mapping with a shared encoder and parameterize each expert as a finite-dimensional linear generator.

\textbf{Latent linear experts.}~
We construct a finite-dimensional Koopman embedding using a shared encoder $\phi_\eta$ and a projector $\xi$.
Given a flow state $x_t$, we compute $h_t=\phi_\eta(x_t)$ and project it into the Koopman latent space as $z_t=\xi(h_t)$.
\methodname{} maintains $K$ experts, each parameterized by a learnable matrix $A^{k}\in\mathbb{R}^{d_z\times d_z}$.
Conditioned on expert $k$, the latent velocity field is:
\begin{equation}
\label{eq:latent_ode}
    \frac{\mathrm{d}z_t}{\mathrm{d}t} = A^{k} z_t.
\end{equation}
Each $A^{k}$ defines a coherent latent transport mode along the flow path.
Its spectrum controls the rotational and contractive components of the latent
velocity over flow time, allowing different experts to specialize in distinct
local transport regimes while retaining linear latent evolution.

\textbf{Expert velocity decoding.}~
To correct the data-space flow, we project the latent velocity $A^{k}z_t$ back to $\mathcal{X}$ using an expert velocity decoder $g_{\psi}$:
\begin{equation}
\label{eq:expert_field}
    v_t^{\psi,k}(x_t, t) = g_{\psi}\!\left(z_t, \, A^{k} z_t\right).
\end{equation}
The decoder maps structured linear dynamics in the Koopman space into a nonlinear data-space residual velocity.

\textbf{Spectral constraints and plasticity.}~
Unconstrained learning of $A^k$ can lead to unstable dynamics and unbounded time-series data. 
To encourage stable trajectories while preserving expressive temporal structure, we enforce dissipativity by requiring
$
\frac{1}{2}\bigl(A^{k}+(A^{k})^\top\bigr)\preceq -\delta I, \delta \ge 0.$
We impose this constraint by parameterizing:
\begin{equation}
\label{eq:stable_param}
A^{k} = \bigl(S^{k}-(S^{k})^\top\bigr) - (R^{k})^\top R^{k} - \delta I,
\end{equation}
where $S^k \in \mathbb{R}^{d_z \times d_z}$ generates the skew-symmetric component that supports oscillatory modes through latent rotations, and $R^k \in \mathbb{R}^{d_z \times d_z}$ parameterizes the dissipative term that damps transient dynamics.
The margin $-\delta I$ ensures spectral dissipativity.
This construction keeps $S^k$ and $R^k$ unconstrained during optimization while guaranteeing stable latent dynamics and preserving broad spectral diversity.

\subsection{WTA Competitive Routing for Flow Matching}
\label{sec:method_wta}

While Koopman experts provide a structured way to model distinct temporal modes, \methodname{} still needs to associate each intermediate flow state with an appropriate expert.
In multi-modal regions, several experts may induce plausible local corrections for the same $x_t$, while soft mixing can average their velocities and reintroduce regression-to-the-mean behavior.
We therefore introduce a Winner-Take-All (WTA) routing rule that assigns each flow state to a dominant expert and encourages mode-specific residual dynamics along the trajectory.

\textbf{Flow routing policy.}~
Along the flow-matching probability path, we use a learnable routing network $\mathcal{R}^\omega$ to assign flow states to experts.
Given flow time $t$ and state $x_t$, the router outputs a categorical distribution over $K$ experts:
\begin{equation}
\label{eq:router_prob}
\pi_t^{\,k} = \mathcal{R}^\omega(t,\phi_\eta(x_t)),
\quad
\text{s.t.}\ \pi_t^{\,k} \ge 0,\ 
\sum\nolimits_{k=1}^{K}\pi_t^{\,k}=1.
\end{equation}
Here, $\pi_t^{\,k}$ denotes the probability of selecting expert $k$ at flow time $t$.
The dependence on $t$ allows routing decisions to adapt to different noise-to-data stages of the transport process.

\textbf{Confidence-aware WTA selection.}~
During training, the target endpoint $x_1$ provides a natural signal for expert assignment.
For each expert $k$, we estimate a candidate endpoint $\hat{x}_1^{\,k}$ using the expert-informed velocity from the current state toward the endpoint.
We then define a confidence-aware score:
\begin{equation}
\label{eq:score_def}
\mathcal{S}^{k}_t(t,x_t,x_1)
=
\|\hat{x}_1^{\,k}-x_1\|_2^2
-
\beta \log\!\big(\pi_t^{\,k}+\varepsilon\big),
\end{equation}
where $\beta$ controls the confidence regularization and $\varepsilon>0$ ensures numerical stability.
The winning expert is selected by a hard WTA rule:
\begin{equation}
\label{eq:wta_winner}
k^*_t
=
\arg\min_{k\in\{1,\ldots,K\}}
\mathcal{S}^{k}_t(t,x_t,x_1).
\end{equation}
So, at each flow time $t$, a single dominant expert provides the primary residual correction for the current state.
This hard assignment avoids soft velocity averaging and promotes specialization across temporal regimes.
The score in Eq.~\eqref{eq:score_def} also stabilizes routing by combining endpoint error with router confidence, reducing noisy early assignments and allowing specialization to develop more smoothly.
\subsection{Learning Mode-Specific Dynamical Experts}
\label{sec:method_optimization}
In this part, we describe the training pipeline of \methodname{}, which learns mode-specific experts to complement the global FM transport in multi-modal regimes.
The global transport is supervised by the CFM loss $\mathcal{L}_{\mathrm{CFM}}$ in Eq.~\eqref{equ:cfm}, which anchors the overall noise-to-data mapping and preserves distributional alignment.
To mitigate estimator-level mean seeking and the resulting mode collapse, we introduce a WTA loss that updates only the selected expert for each $(t,x_t)$.
Given the winner $k_t^*$ from Eq.~\eqref{eq:wta_winner}, we optimize
\begin{equation}
\label{eq:l_wta}
\mathcal{L}_{\mathrm{WTA}}
=
\mathbb{E}_{t,x_0,x_1}\!\left[
\lambda_t\,\mathcal{S}^{k^*_t}_t(t,x_t,x_1)
\right],
\end{equation}
where $\lambda_t$ is a time-dependent weighting schedule along the flow trajectory.
Gradients from $\mathcal{L}_{\mathrm{WTA}}$ are explicitly masked for all non-selected experts, so specialization is driven only by the expert that best explains the current sample.

To avoid routing collapse and encourage diverse expert usage, we add a load-balancing regularizer:
\begin{equation}
\label{eq:l_bal}
\bar{\pi}
=
\frac{1}{B}\sum_{i=1}^{B}
\pi_{t^{(i)}}^{(i)},
\qquad
\mathcal{L}_{\mathrm{BAL}}
=
\mathrm{KL}\!\left(u \,\middle\|\, \bar{\pi}\right),
\end{equation}
where $u=(1/K,\ldots,1/K)$ is the uniform prior and $B$ is the batch size.
Since hard routing can overuse a few experts early in training, $\mathcal{L}_{\mathrm{BAL}}$ penalizes batch-level imbalance, mitigating dead experts and promoting sufficient specialization signals for all experts.

Thus, the overall objective is 
\begin{equation}
\small
\label{eq:total_loss}
\mathcal{L}_{\mathrm{total}}
=
\mathcal{L}_{\mathrm{CFM}}(\theta)
+
\alpha_{W}\,\mathcal{L}_{\mathrm{WTA}}(\eta,\psi,\xi,\omega,A)
+
\alpha_{B}\,\mathcal{L}_{\mathrm{BAL}}(\omega).
\end{equation}
Here, $\mathcal{L}_{\mathrm{CFM}}$ learns the shared global transport estimator, $\mathcal{L}_{\mathrm{WTA}}$ promotes mode-specific residual corrections through hard expert assignment, and $\mathcal{L}_{\mathrm{BAL}}$ prevents degenerate routing.

\begin{table*}[!t]
  \centering
  \scriptsize  % 使用scriptsize替代tiny，更清晰可读
  \caption{Results of all methods on all datasets.}\label{table:unconditional_time_series}
  \setlength{\tabcolsep}{9pt}  % 减少列间距使表格更紧凑
  \renewcommand{\arraystretch}{0.95}  % 减少行距
  \resizebox{1\textwidth}{!}{
    \begin{tabular}{>{\centering\arraybackslash}m{2.2cm}|c|cccccc}
    \toprule
    {Metric} & {Methods} & {Sines} & {Stocks} & {ETTh} & {MuJoCo} & {Energy} & {fMRI} \\
    \midrule
    \multirow{8}{*}{{Context-FID  Score $\downarrow$}} & \cellcolor[rgb]{0.88,0.94,0.99} Ours & \cellcolor[rgb]{0.88,0.94,0.99}\textbf{0.003±.000} & \cellcolor[rgb]{0.88,0.94,0.99}\underline{0.027±.008}  & \cellcolor[rgb]{0.88,0.94,0.99}\textbf{0.015±.001} & \cellcolor[rgb]{0.88,0.94,0.99}\textbf{0.006±.005} & \cellcolor[rgb]{0.88,0.94,0.99}\textbf{0.022±.002} & \cellcolor[rgb]{0.88,0.94,0.99}0.155±.002 \\
    & SMF~\cite{kim2025sequence} &
    \underline{0.005±.001} & \textbf{0.023±.003} & \underline{0.059±.007} & {0.019±.002} & \underline{0.049±.007} & \underline{0.116±.006} \\
    & Diffusion-TS~\cite{yuan2024diffusion} &
    0.014±.001 & 0.208±.052 & {0.132±.010} & \underline{0.015±.001} & 0.092±.020 & \textbf{0.108±.006} \\
    & DiffTime~\cite{coletta2023constrained} &
    {0.008±.002} & 0.236±.074 & 0.299±.044 & 0.188±.028 & 0.289±.045 & 0.324±.045 \\
    & Diffwave~\cite{kong2020diffwave} &
    0.014±.002 & 0.232±.032 & 0.873±.061 & 0.393±.041 & 1.031±.131 & 0.244±.018 \\
    & TimeGAN~\cite{yoon2019time} &
    0.101±.014 & 0.105±.025 & 0.300±.013 & 0.565±.028 & 0.742±.133 & 1.292±.218 \\
    & TimeVAE~\cite{desai2021timevae} &
    0.307±.060 & 0.215±.035 & 0.805±.186 & 0.251±.015 & 1.631±.142 & 14.449±.969 \\
    & Cot-GAN~\cite{xu2020cot} &
    1.337±.068 & 0.408±.086 & 0.980±.071 & 1.094±.079 & 1.039±.028 & 7.813±.550 \\
    \midrule
    \multirow{8}{*}{{Correlational Score $\downarrow$}} & \cellcolor[rgb]{0.88,0.94,0.99} Ours & \cellcolor[rgb]{0.88,0.94,0.99}\textbf{0.017±.004} & \cellcolor[rgb]{0.88,0.94,0.99}\underline{0.007±.003} & \cellcolor[rgb]{0.88,0.94,0.99}\textbf{0.022±.010} & \cellcolor[rgb]{0.88,0.94,0.99}\underline{0.203±.030} & \cellcolor[rgb]{0.88,0.94,0.99}\textbf{0.495±.033} & \cellcolor[rgb]{0.88,0.94,0.99}\underline{0.898±.025} \\
    & SMF~\cite{kim2025sequence} &
    {0.027±.012} & {0.010±.007} & \underline{0.034±.006} & 0.210±.032 & 0.900±.277 & \textbf{0.774±.017} \\
    & Diffusion-TS~\cite{yuan2024diffusion} &
    {0.060±.008} & {0.013±.009} & 0.059±.008 & \textbf{0.188±.035} & \underline{0.859±.183} & 1.416±.028 \\
    & DiffTime~\cite{coletta2023constrained} &
    \textbf{0.017±.004} & \textbf{0.006±.002} & 0.067±.005 & 0.238±.031 & 1.158±.095 & 1.701±.048 \\
    & Diffwave~\cite{kong2020diffwave} &
    \underline{0.022±.005} & 0.030±.020 & 0.175±.006 & 0.479±.018 & 5.001±.154 & 3.927±.049 \\
    & TimeGAN~\cite{yoon2019time} &
    0.045±.010 & 0.063±.005 & 0.210±.006 & 0.886±.035 & 4.010±.104 & 23.502±.039 \\
    & TimeVAE~\cite{desai2021timevae} &
    0.131±.010 & 0.095±.008 & 0.111±.020 & 0.388±.041 & 1.728±.226 & 17.296±.526 \\
    & Cot-GAN~\cite{xu2020cot} &
    0.049±.010 & 0.087±.004 & 0.249±.009 & 1.042±.007 & 3.164±.061 & 26.824±.449 \\
    \midrule
    \multirow{9}{*}{{Discriminative Score $\downarrow$}} & \cellcolor[rgb]{0.88,0.94,0.99} Ours & \cellcolor[rgb]{0.88,0.94,0.99}\textbf{0.004±.003} & \cellcolor[rgb]{0.88,0.94,0.99}\underline{0.030±.004} & \cellcolor[rgb]{0.88,0.94,0.99}\textbf{0.005±.004} & \cellcolor[rgb]{0.88,0.94,0.99}\textbf{0.004±.008} & \cellcolor[rgb]{0.88,0.94,0.99}\textbf{0.039±.008} & \cellcolor[rgb]{0.88,0.94,0.99}\textbf{0.129±.053} \\
    & SMF~\cite{kim2025sequence} &
    \underline{0.005±.006} & \textbf{0.027±.014} & \underline{0.017±.009} & \underline{0.005±.004} & \underline{0.150±.018} & \underline{0.136±.139} \\
    & Diffusion-TS~\cite{yuan2024diffusion} &
    0.006±.005 & {0.086±.024} & 0.077±.010 & 0.012±.006 & 0.139±.012 & 0.200±.083 \\
    & DiffTime~\cite{coletta2023constrained} &
    0.013±.006 & 0.107±.016 & 0.110±.007 & 0.154±.045 & 0.485±.002 & 0.245±.051 \\
    & Diffwave~\cite{kong2020diffwave} &
    0.017±.008 & 0.232±.061 & 0.190±.008 & 0.203±.096 & 0.473±.004 & 0.402±.029 \\
    & TimeGAN~\cite{yoon2019time} &
    {0.011±.008} & 0.102±.021 & 0.114±.055 & 0.238±.068 & 0.236±.012 & 0.484±.042 \\
    & TimeVAE~\cite{desai2021timevae} &
    0.041±.044 & 0.145±.120 & 0.209±.058 & 0.230±.102 & 0.499±.000 & 0.476±.044 \\
    & Cot-GAN~\cite{xu2020cot} &
    0.254±.137 & 0.230±.016 & 0.325±.099 & 0.426±.022 & 0.498±.002 & 0.492±.018 \\
    & RNN-AR~\cite{yoon2019time} &
    0.495±.001 & 0.226±.035 & - & - & 0.483±.004 & - \\
    \midrule
    \multirow{10}{*}{{Predictive Score $\downarrow$}} & \cellcolor[rgb]{0.88,0.94,0.99} Ours & \cellcolor[rgb]{0.88,0.94,0.99}\textbf{0.093±.000} & \cellcolor[rgb]{0.88,0.94,0.99}\textbf{0.037±.000} & \cellcolor[rgb]{0.88,0.94,0.99}0.122±.003 & \cellcolor[rgb]{0.88,0.94,0.99}\textbf{0.006±.003} & \cellcolor[rgb]{0.88,0.94,0.99}\underline{0.251±.003} & \cellcolor[rgb]{0.88,0.94,0.99}\textbf{0.100±.000} \\
    & SMF~\cite{kim2025sequence} &
    \textbf{0.093±.000} & \textbf{0.037±.000} & 0.123±.005 & 0.008±.001 & \textbf{0.193±.045} & \textbf{0.100±.000} \\
    & Diffusion-TS~\cite{yuan2024diffusion} &
    \underline{0.095±.000} & \textbf{0.037±.000} & \textbf{0.118±.008} & \underline{0.007±.001} & \underline{0.251±.000} & \textbf{0.100±.000} \\
    & DiffTime~\cite{coletta2023constrained} &
    \textbf{0.093±.000} & \underline{0.038±.001} & \underline{0.121±.004} & 0.010±.001 & 0.252±.000 & \textbf{0.100±.000} \\
    & Diffwave~\cite{kong2020diffwave} &
    \textbf{0.093±.000} & 0.047±.000 & 0.130±.001 & 0.013±.000 & \underline{0.251±.000} & \underline{0.101±.000} \\
    & TimeGAN~\cite{yoon2019time} &
    \textbf{0.093±.019} & \underline{0.038±.001} & 0.124±.001 & 0.025±.003 & 0.273±.004 & 0.126±.002 \\
    & TimeVAE~\cite{desai2021timevae} &
    \textbf{0.093±.000} & 0.039±.000 & 0.126±.004 & 0.012±.002 & 0.292±.000 & 0.113±.003 \\
    & Cot-GAN~\cite{xu2020cot} &
    0.100±.000 & 0.047±.001 & 0.129±.000 & 0.068±.009 & 0.259±.000 & 0.185±.003 \\
    & RNN-AR~\cite{yoon2019time} &
    0.150±.022 & \underline{0.038±.001} & - & - & 0.315±.005 & - \\
    \cmidrule{2-8}
    & Original & {0.094±.001} & {0.036±.001} & {0.121±.005} & {0.007±.001} & {0.250±.003} & {0.090±.001} \\
    \bottomrule
    \end{tabular}}
  \vspace{-0.2cm}
\end{table*}

\subsection{Sampling with Expert-Informed Residual Dynamics}
\label{sec:method_sampling}
During inference, \methodname{} follows the globally learned FM transport field while the router selects a dominant expert $k_t^*=\arg\max_k \pi_t^k$ for each state. The selected expert provides a residual correction that steers the trajectory toward mode-consistent local evolution:
\begin{equation}
\label{eq:sampling_update}
x_{t+\Delta t} = x_t + \Big( v_t^{\theta}(x_t, t) + \gamma \lambda_t v_t^{\psi, k^*_t}(x_t, t) \Big)\Delta t,
\end{equation}
This update preserves spectral patterns that tend to be lost under global averaging, yielding samples that are statistically consistent and dynamically expressive.

\section{Empirical Evaluation}
\label{sec:experiments}
We evaluate \methodname{} on real-world and synthetic datasets for unconditional and conditional generation, measuring sample quality, spectral diversity, and multi-scale temporal preservation. We also run ablations to verify each component and examine how experts partition the data manifold.

\textbf{Datasets \& Metrics.} Following \cite{yoon2019time,yuan2024diffusion}, we evaluate our model on four real-world benchmarks (Stocks, ETTh, Energy, and fMRI) and two synthetic datasets (Sines and MuJoCo), covering varying degrees of dynamical complexity.
For evaluation, we adopt Context-FID \cite{jeha2022psa} and the Correlational Score \cite{liao2020conditional} to measure distributional alignment and assess temporal dependencies.
We additionally use the Discriminative Score \cite{yoon2019time} and Predictive Score \cite{yoon2019time} to evaluate statistical fidelity.

\subsection{Evaluation of Mode Diversity}

\begin{wrapfigure}[19]{r}{0.56\textwidth}
    \vspace{-1.5cm}
    \centering
    \includegraphics[width=\linewidth]{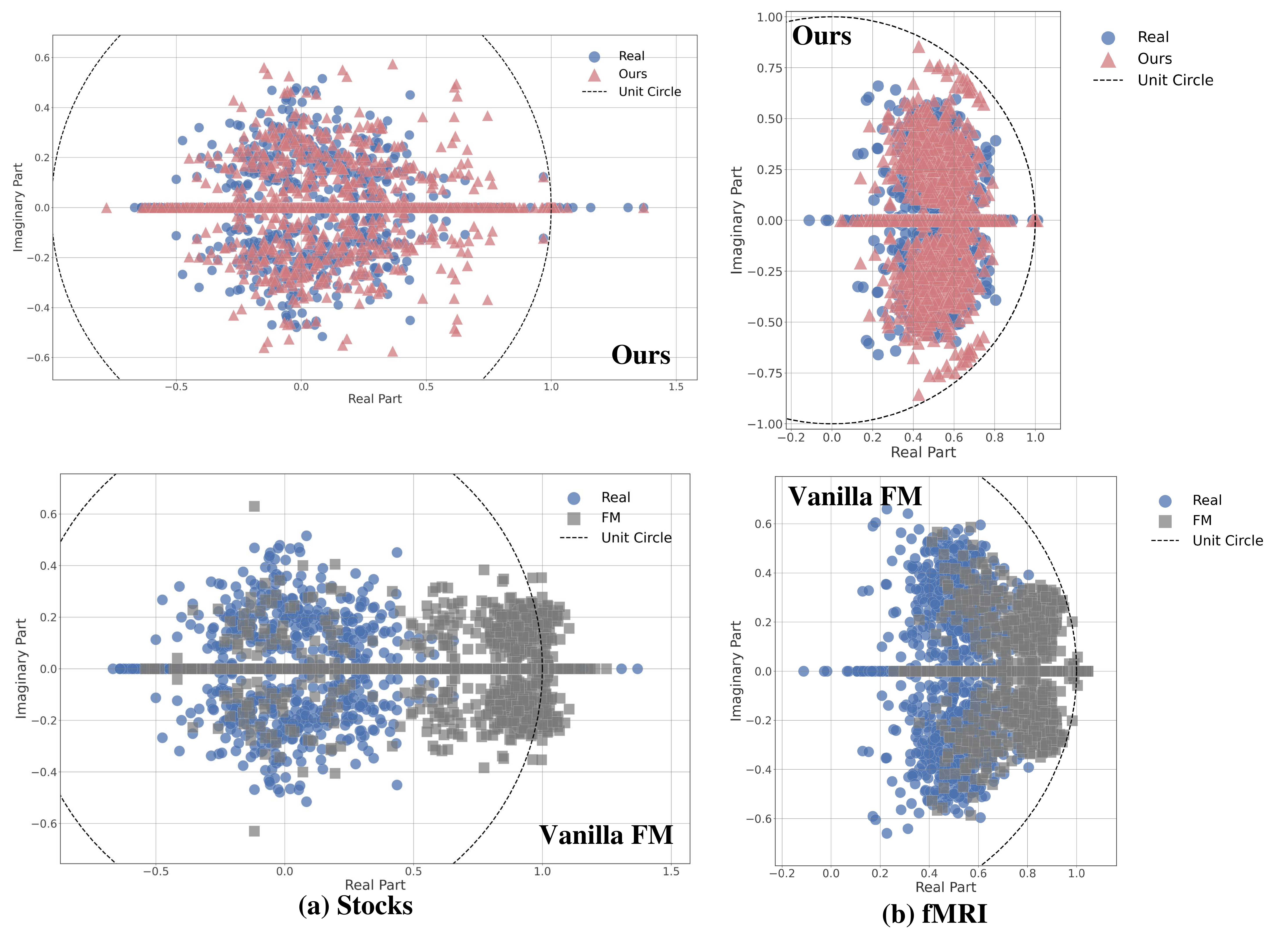}
    \caption{Comparison of DMD eigenvalues between real and generated time series for our \methodname{} and vanilla FM. The imaginary axis represents the oscillation frequency (larger absolute values indicate faster oscillations), while the distance from the origin indicates system stability.}
    \label{fig:dmd}

\end{wrapfigure}

We first examine  mode collapse by comparing \methodname{} with vanilla Flow Matching.
Figure~\ref{fig:dmd} illustrates  the DMD eigenvalue distributions on the Stocks and fMRI datasets.
The imaginary axis reflects oscillation frequency, and a wider spread indicates richer temporal dynamics across slow and fast variations.
The real data (blue circles) exhibit broad spectral support and clear geometric structure, suggesting complex multi-scale dynamics rather than a single dominant time scale.

Vanilla FM shows evident spectral contraction: its eigenvalues (gray squares) cluster near the real axis and span a narrower range along the imaginary direction.
This pattern indicates that the learned single-field estimator favors a few slow modes, while attenuating high-frequency and transient components after ODE integration.
Such contraction is consistent with the estimator-level regression-to-the-mean behavior discussed in Sec.~\ref{sec:prelim}, and results in overly smooth generated trajectories.
But \methodname{} (pink triangles) achieves much denser spectral overlap with the real data.
By using expert-informed residual dynamics to complement the global transport estimator, \methodname{} reduces mean-seeking effects and preserves a wider range of oscillatory modes.
Consequently, the generated signals retain broader spectral diversity and multi-scale temporal structure, better matching the dynamical patterns of the empirical distribution.
\begin{table*}[!t]
  \centering
  \scriptsize
  \caption{Performance under different training data sizes on the Stocks dataset.}
  \label{tab:drop_metrics}
  \setlength{\tabcolsep}{6pt}
  \renewcommand{\arraystretch}{0.95}
  \resizebox{0.90\textwidth}{!}{
  \begin{tabular}{@{}c|c|cccc@{}}
  \hline
  {Metric} & {Methods} & {20\%} & {40\%} & {60\%} & {80\%} \\
  \hline
  \multirow{3}{*}{\shortstack{Context-FID\\Score $\downarrow$}} 
  & \cellcolor[rgb]{0.88,0.94,0.99} Ours 
  & \cellcolor[rgb]{0.88,0.94,0.99}\textbf{1.920$\pm$.574} 
  & \cellcolor[rgb]{0.88,0.94,0.99}\textbf{1.890$\pm$.251} 
  & \cellcolor[rgb]{0.88,0.94,0.99}\textbf{1.789$\pm$.437} 
  & \cellcolor[rgb]{0.88,0.94,0.99}\textbf{0.824$\pm$.434} \\
  & {SMF} 
  & 6.406$\pm$.222 
  & 3.900$\pm$.455 
  & \underline{2.613$\pm$.567} 
  & {0.887$\pm$.054} \\
  & {Diffusion-TS} 
  & \underline{5.197$\pm$.583} 
  & \underline{3.469$\pm$.389} 
  & 2.854$\pm$.583 
  & \underline{0.860$\pm$.204} \\
  \hline
  \multirow{3}{*}{\shortstack{Discriminative\\Score $\downarrow$}} 
  & \cellcolor[rgb]{0.88,0.94,0.99} Ours 
  & \cellcolor[rgb]{0.88,0.94,0.99}\textbf{0.303$\pm$.004} 
  & \cellcolor[rgb]{0.88,0.94,0.99}\textbf{0.242$\pm$.012} 
  & \cellcolor[rgb]{0.88,0.94,0.99}\textbf{0.194$\pm$.005} 
  & \cellcolor[rgb]{0.88,0.94,0.99}\textbf{0.084$\pm$.005} \\
  & {SMF} 
  & \underline{0.347$\pm$.008} 
  & \underline{0.259$\pm$.014} 
  & \underline{0.202$\pm$.006} 
  & \underline{0.097$\pm$.006} \\
  & {Diffusion-TS} 
  & 0.351$\pm$.006 
  & 0.267$\pm$.010 
  & 0.221$\pm$.008 
  & 0.158$\pm$.018 \\
  \hline
  \end{tabular}
  }
  \vspace{-0.2cm}
\end{table*}

\begin{figure*}[t]
    % \vspace{-10pt}
    % \hspace*{-5pt}
    \centering
    \includegraphics[width=1\linewidth]{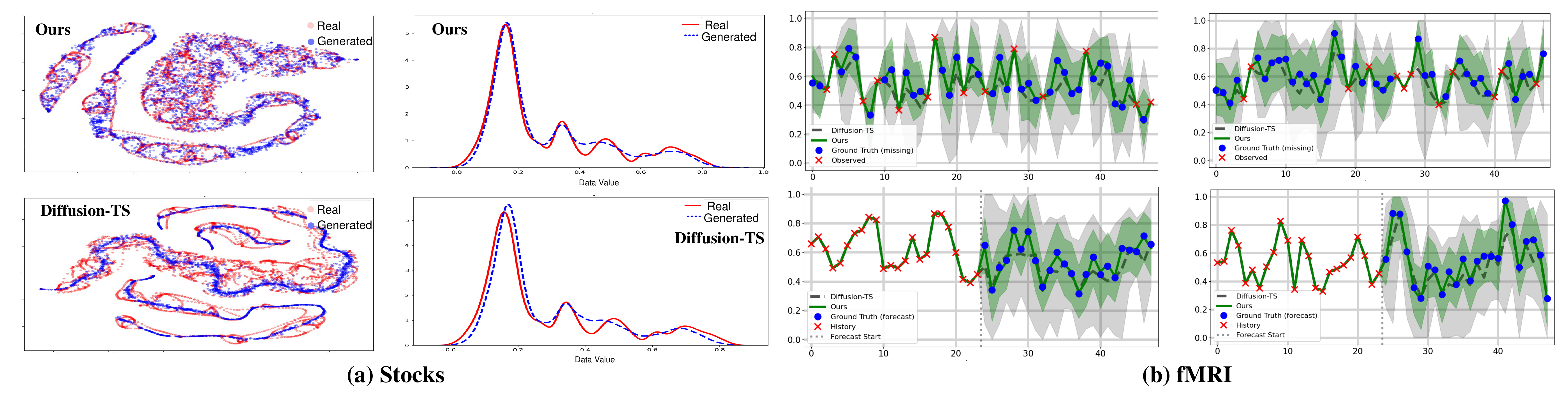}
    
    \caption{Comparison of unconditional time-series generation on the Stocks dataset (left) and imputation (top row) and forecasting (bottom row) on the fMRI dataset (right) between our \methodname{} and Diffusion-TS.}
    \label{fig:visualization}
\vspace{-0.6cm}
\end{figure*}

\subsection{Unconditional Time Series Generation}

We evaluate \methodname{} on unconditional time-series generation under comparable parameter budgets for all baselines.
The compared methods include generative adversarial networks (GANs)~\cite{yoon2019time,xu2020cot}, variational autoencoders (VAEs)~\cite{desai2021timevae}, diffusion-based models~\cite{kong2020diffwave,yuan2024diffusion,coletta2023constrained}, and FM-based models~\cite{kim2025sequence}.

Table~\ref{table:unconditional_time_series} reports generation performance across diverse datasets.
\methodname{} achieves the best results in 16 out of 24 cases, outperforming recent baselines such as SMF and Diffusion-TS.
These results are consistent with the spectral analysis above: by using routed residual experts to complement the global estimator, \methodname{} alleviates estimator-level mean seeking and better preserves multi-scale temporal modes.
The benefit is especially clear on complex datasets such as Energy, where \methodname{} reduces Context-FID by over 55\% and Discriminative Score by about 74\% compared with the second-best method.
On the high-dimensional fMRI dataset, \methodname{} also achieves the lowest Discriminative Score, indicating that it better preserves rhythmic and structured dynamics that are often attenuated by over-smoothed learned flows.
Overall, the results show that \methodname{} improves both statistical fidelity and temporal realism by maintaining a broader spectral range and capturing multi-scale structures in time-series data.

We further visualize generation quality using t-SNE projections and kernel density estimation (KDE).
As shown in Figure~\ref{fig:visualization}(a), samples generated by \methodname{} overlap closely with real data in both views.
In the t-SNE plot, generated trajectories follow the global structure of real samples and cover both central regions and more complex outer areas.
In contrast, Diffusion-TS leaves visible gaps in the sample space, suggesting partial mode dropping.
The KDE plots show a similar trend: \methodname{} closely matches the real density curves and captures the multiple peaks and sharp variations in the Stocks dataset.
Our appendix further shows that \methodname{} achieves this performance with high sampling efficiency.

\textbf{Small-Scale Settings.}~
We also evaluate \methodname{} under severe data scarcity by training on reduced subsets of the training data.
As shown in Tab.~\ref{tab:drop_metrics}, \methodname{} remains stable across all low-data settings and achieves the best Context-FID and Discriminative Scores among competing baselines.
With only 20\% of the training data, \methodname{} still yields clear gains, improving Context-FID by 63.1\% and Discriminative Score by 12.7\%.
We attribute this robustness to expert-specific residual modeling: by assigning distinct temporal regimes to specialized experts, \methodname{} preserves subtle temporal variations that are easily averaged out in low-data regimes.
As a result, it maintains fine-grained dynamics and high generation quality even with substantially fewer training examples.

% We compare against strong baselines, including Diffusion-TS, which extends naturally to conditional settings.
\subsection{Conditional Time Series Generation}
Beyond unconditional generation, we extend \methodname{} to conditional tasks, including forecasting and imputation, using flow-matching guidance~\cite{feng2025guidance}.
Specifically, we add a guidance gradient that enforces consistency with observed values and use it as an additional correction to the sampling velocity.
As shown in Figure~\ref{fig:visualization}(b), \methodname{} better follows sharp transitions and sudden spikes where Diffusion-TS often falls short.
This advantage comes from the Koopman-inspired experts, which provide structured temporal inductive biases, and the WTA routing mechanism, which promotes specialization over fine-grained temporal modes.
As a result, the model can recover more faithful trajectories from limited local observations.

\subsection{Ablation Study}

\begin{wraptable}[12]{r}{0.56\textwidth}
\vspace{-1.7cm}
\centering
\scriptsize
\setlength{\tabcolsep}{3pt}
\renewcommand{\arraystretch}{0.98}
\caption{Ablation study on the model design.}
\label{tab:ablation_Stocks_fMRI}
\resizebox{\linewidth}{!}{
\begin{tabular}{l|lcc}
\toprule
Metric & Variant & Stocks & fMRI \\
\midrule
\multirow[c]{7}{*}{\makecell[c]{Context-FID\\Score $\downarrow$}}
 & \cellcolor[rgb]{0.88,0.94,0.99} Ours 
 & \cellcolor[rgb]{0.88,0.94,0.99}\textbf{0.027$\pm$.008} 
 & \cellcolor[rgb]{0.88,0.94,0.99}\textbf{0.155$\pm$.002} \\
& Vanilla FM & 0.116$\pm$.012 & 0.324$\pm$.030 \\
\cmidrule(lr){2-4}
& \textit{Training}~w/o $\mathcal{L}_{\mathrm{WTA}}$ & 0.107$\pm$.011 & 0.294$\pm$.018 \\
& \textit{Training}~w/o $\mathcal{L}_{\mathrm{BAL}}$ & 0.088$\pm$.015 & 0.178$\pm$.007 \\
& \textit{Training}~$\beta=0$ & 0.095$\pm$.010 & 0.248$\pm$.019 \\
\cmidrule(lr){2-4}
& \textit{Sampling}~$\gamma=0$ & 0.091$\pm$.010 & 0.185$\pm$.014 \\
\midrule
\multirow[c]{7}{*}{\makecell[c]{Discriminative\\Score $\downarrow$}}
& \cellcolor[rgb]{0.88,0.94,0.99} Ours 
& \cellcolor[rgb]{0.88,0.94,0.99}\textbf{0.030$\pm$.004} 
& \cellcolor[rgb]{0.88,0.94,0.99}\textbf{0.129$\pm$.053} \\
& Vanilla FM & 0.186$\pm$.005 & 0.245$\pm$.050 \\
\cmidrule(lr){2-4}
& \textit{Training}~w/o $\mathcal{L}_{\mathrm{WTA}}$ & 0.112$\pm$.014 & 0.402$\pm$.029 \\
& \textit{Training}~w/o $\mathcal{L}_{\mathrm{BAL}}$ & 0.106$\pm$.045 & 0.218$\pm$.020 \\
& \textit{Training}~$\beta=0$ & 0.102$\pm$.074 & 0.229$\pm$.058 \\
\cmidrule(lr){2-4}
& \textit{Sampling}~$\gamma=0$ & 0.075$\pm$.012 & 0.275$\pm$.029 \\
\bottomrule
\end{tabular}
}
\vspace{0.2cm}
\end{wraptable}

Table~\ref{tab:ablation_Stocks_fMRI} further shows the contribution of each component.
\methodname{} consistently outperforms vanilla FM, confirming the benefit of augmenting the global transport estimator with mode-specific residual dynamics.
Among all components, removing the WTA objective $\mathcal{L}_{\mathrm{WTA}}$ causes the largest performance drop, especially on the high-dimensional fMRI dataset.
Without competitive specialization, experts are less clearly separated across temporal regimes, and the learned dynamics become more mean-seeking, producing smoother trajectories that miss fine-grained structures.
The load-balancing term $\mathcal{L}_{\mathrm{BAL}}$ and confidence weight $\beta$ further improve routing behavior by preventing a small subset of experts from dominating.
This is particularly useful for datasets such as Stocks, where temporal patterns are noisy and shift over time.
By encouraging broader expert utilization, these components help retain both frequent and less common temporal modes.
Finally, performance degrades when $\gamma=0$, indicating that expert-informed residual dynamics during sampling are essential for generation quality.
Without this residual correction, the sampler relies mainly on the global estimator and becomes less adaptive to diverse temporal regimes, losing important local variations.
This supports our design principle: combining stable global transport with competitively routed residual experts mitigates estimator-level averaging and better captures complex temporal structure.

\subsection{Analysis of Mode-Specific Experts}
\label{sec:specialization}

\begin{wrapfigure}[22]{r}{0.52\textwidth}
    \vspace{-1.1cm}
    \centering
    \includegraphics[width=\linewidth]{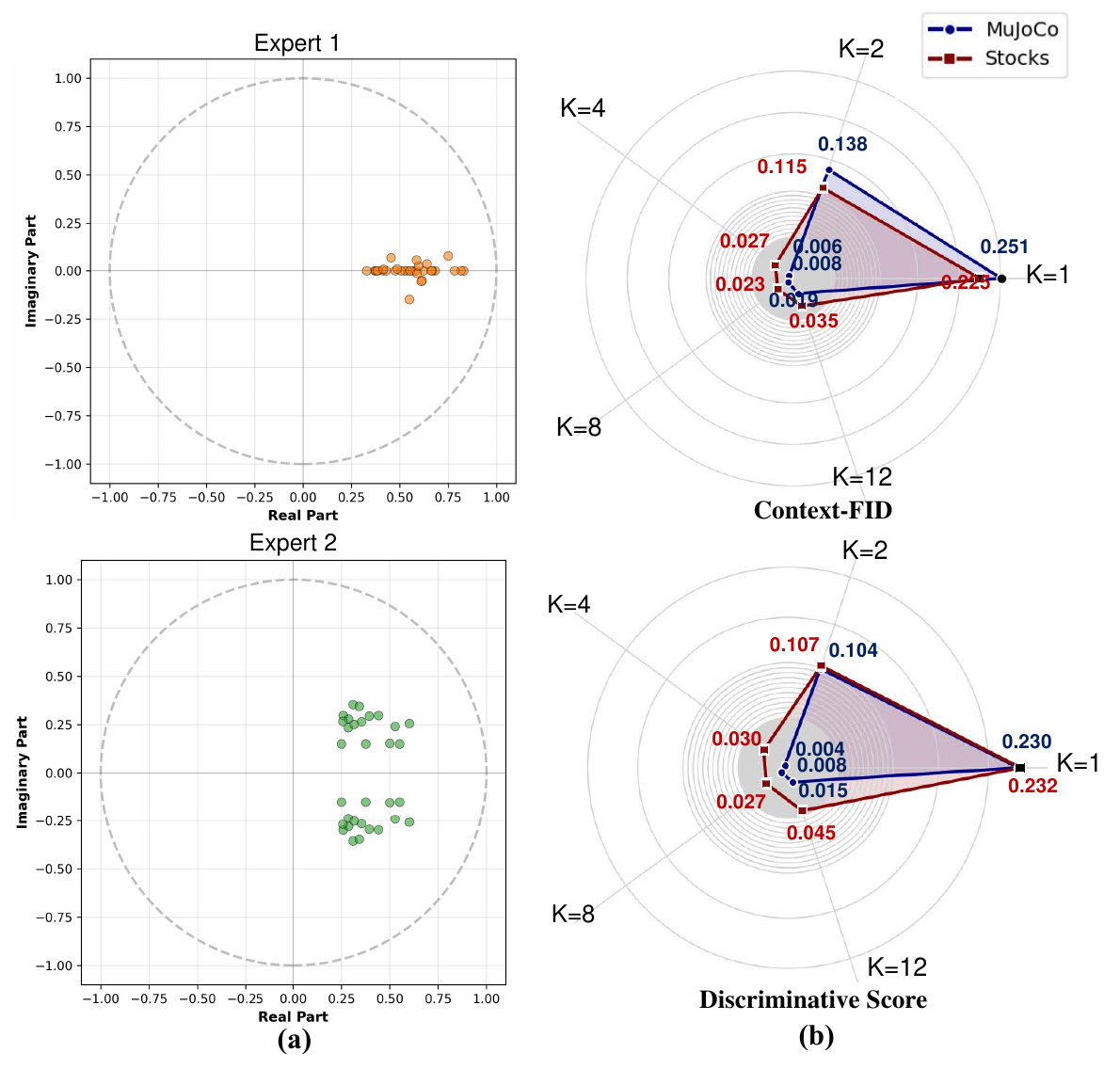}
    \caption{(a) DMD eigenvalues for Expert 1 and Expert 2 on fMRI. 
    (b) Effect of the number of experts on the unconditional task for the MuJoCo and Stocks datasets. Lower is better.}
    \label{fig:exp}
    % \vspace{1cm}
\end{wrapfigure}

To better understand how \methodname{} reproduces the temporal modes in Figure~\ref{fig:dmd}, we examine the DMD eigenvalue spectra of the trained fMRI experts in Figure~\ref{fig:exp}(a).
The expert spectra show a clear division of labor that mirrors the main structures of the real spectrum.
Expert 1 mainly produces eigenvalues close to the real axis, corresponding to non-oscillatory components such as slow drifts and smooth trend-like dynamics.
These modes match the band near the real axis in Figure~\ref{fig:dmd}.
In contrast, Expert 2 focuses on oscillatory components with larger imaginary parts, forming conjugate clusters above and below the real axis.
This structure matches the symmetric clouds in the real DMD spectrum and reflects rhythmic fluctuations in fMRI signals.
These results explain why \methodname{} improves spectral fidelity: rather than forcing a single learned estimator to average incompatible dynamics, the WTA routing mechanism assigns different temporal regimes to specialized experts.
As a result, the generated sequences preserve diverse dynamics and yield spectra that better match the empirical distribution.

We further analyze the effect of the number of experts $K$ in Figure~\ref{fig:exp}(b).
Across datasets, $K=1$ performs the worst, confirming that a single expert lacks sufficient capacity to capture heterogeneous temporal structures.
Using $K=4$ provides consistently strong performance and serves as our default setting, while the optimal capacity still depends on data complexity.
For Stocks, increasing $K$ to 8 further improves performance, suggesting that additional experts help model its diverse and shifting temporal patterns.
For MuJoCo, performance improves with a moderate number of experts but slightly drops when too many are used, likely because the expert pool exceeds the intrinsic complexity of the data and weakens clear specialization.

\section{Conclusion}

In this paper, we introduce PrismFlow, a new FM method with residual dynamical experts for time-series generation. The method is motivated by a key limitation of learned single-field FM models: when time-series data contain multiple competing temporal regimes, a single learned transport field may suffer from estimator-level averaging, which smooths distinct dynamics and can lead to mode collapse. PrismFlow addresses this problem by decomposing the transport process into a stable global backbone and hard-routed Koopman-inspired residual experts. The global field captures shared temporal evolution, while the experts model regime-specific residual dynamics, such as oscillatory and high-frequency patterns.
Empirical results on various generation tasks show that PrismFlow achieves strong performance and effectively mitigates mode collapse by preserving diverse spectral patterns, unlocking a new paradigm for robust generative modeling.
% This work advances time-series generation by demonstrating that expert-specialized transport fields provide a principled and effective way to model complex and diverse temporal distributions.

\bibliographystyle{unsrtnat}
\bibliography{ref}


%%%%%%%%%%%%%%%%%%%%%%%%%%%%%%%%%%%%%%%%%%%%%%%%%%%%%%%%%%%%

% \clearpage

% \input{checklist_our.tex}
\end{document}